\documentclass[lettersize,journal]{IEEEtran}
\usepackage{amsmath,amsfonts}
\usepackage{algorithmic}
\usepackage{array}
\usepackage[caption=false,font=normalsize,labelfont=sf,textfont=sf]{subfig}
\usepackage{textcomp}
\usepackage{stfloats}
\usepackage{url}
\usepackage{verbatim}
\usepackage{graphicx}
\usepackage{booktabs}
\usepackage{multirow}
\usepackage{makecell}
\usepackage{subcaption}
\usepackage{algorithm}
\usepackage{hyperref}
\usepackage{extra_style}

\def\BibTeX{{\rm B\kern-.05em{\sc i\kern-.025em b}\kern-.08em
    T\kern-.1667em\lower.7ex\hbox{E}\kern-.125emX}}
\usepackage{balance}
\begin{document}
\title{Deep Understanding of Sign Language\\for Sign to Subtitle Alignment}
\author{\textbf{Youngjoon Jang}$^{\dagger}$ \quad \textbf{Jeongsoo Choi}$^{\dagger}$ \quad \textbf{Junseok Ahn} \quad \textbf{Joon Son Chung}\thanks{$^{\dagger}$Both authors contributed equally to this work.}
\vspace{1mm}
\\Korea Advanced Institute of Science and Technology
}

\maketitle

\begin{figure*}[!t]
\centering
  \includegraphics[width=0.95\textwidth]{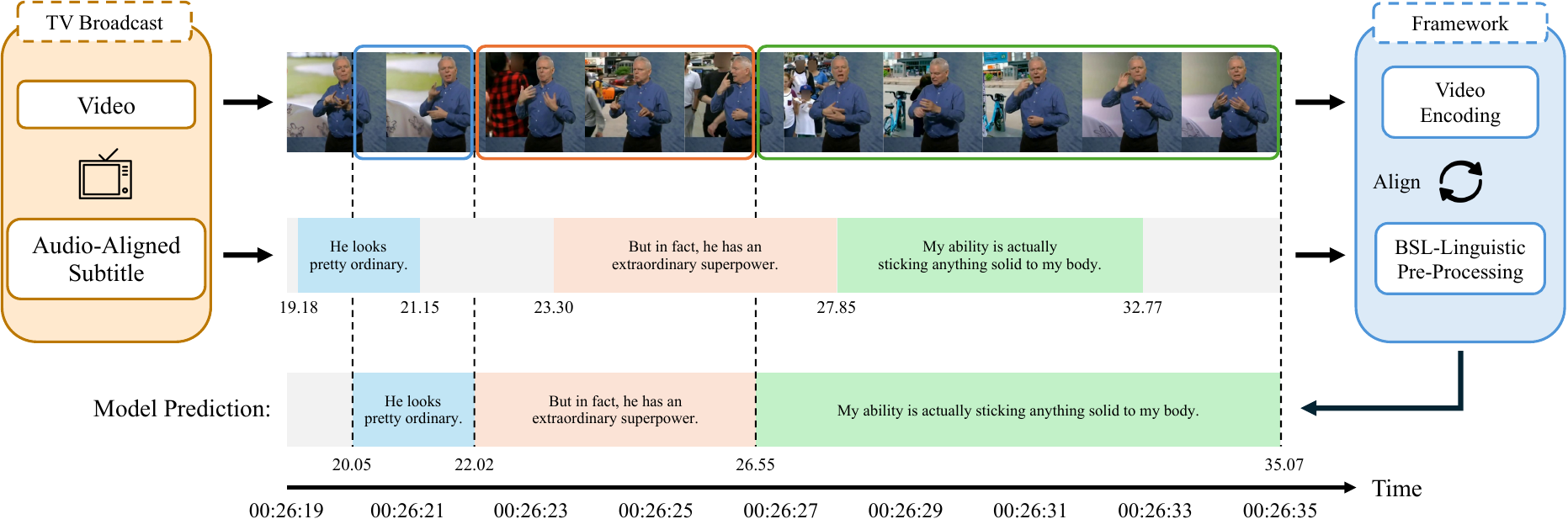}
  \vspace{-1mm}
  \caption{This work aims to align subtitles with continuous signing in sign language interpreted TV broadcast data by leveraging the grammatical systems of British Sign Language. Using two different modalities—video and audio-aligned subtitles—our framework encodes visual features and pre-processes the input query text based on the linguistics of BSL. The output consists of time segments that indicate the points in time when the sign language corresponding to the text is uttered.}
  \label{fig:teaser}
  \vspace{-3mm}
\end{figure*}

\begin{abstract}
The objective of this work is to align asynchronous subtitles in sign language videos with limited labelled data. 
To achieve this goal, we propose a novel framework with the following contributions: 
(1) we leverage fundamental grammatical rules of British Sign Language (BSL) to pre-process the input subtitles, 
(2) we design a selective alignment loss to optimise the model for predicting the temporal location of signs only when the queried sign actually occurs in a scene, and 
(3) we conduct self-training with refined pseudo-labels which are more accurate than the heuristic audio-aligned labels.
From this, our model not only better understands the correlation between the text and the signs, but also holds potential for application in the translation of sign languages, particularly in scenarios where manual labelling of large-scale sign data is impractical or challenging. 
Extensive experimental results demonstrate that our approach achieves state-of-the-art results, surpassing previous baselines by substantial margins in terms of both frame-level accuracy and F1-score. This highlights the effectiveness and practicality of our framework in advancing the field of sign language video alignment and translation. The code is available at \href{https://github.com/art-jang/sign-to-subtitle}{https://github.com/art-jang/sign-to-subtitle}.
\end{abstract}


\section{Introduction}
\IEEEPARstart{S}{ign} language plays a crucial role in enabling communication for deaf individuals. To foster communication between hearing and deaf individuals, it is important to explore applications and technologies that can interpret linguistic meanings from sign language videos.
With advancements in deep learning, recent works~\cite{hu2023continuous, chen2022simple, chen2022two, zuo2022c2slr, kan2022sign, camgoz2020multi} have demonstrated promising results in automatic sign language recognition and translation. However, these achievements have been restricted to environments where continuous signing is manually segmented into individual clips. A key challenge in scaling up such tasks is the difficulty in acquiring large-scale sign language training data consisting of pairs of sentence-level text labels and videos.

In response to this, recent research has utilised sign language broadcasted on TV, which provides videos with continuous signing alongside the transcriptions of the corresponding spoken audio, to construct large-scale datasets~\cite{camgoz2021content4all,Albanie2021bobsl}. While these datasets represent a significant improvement in scale compared to the previous benchmarks~\cite{camgoz2018neural,koller2015continuous,zhou2021improving}, their supervision on signed content is limited in two  aspects. 
First, it is noisy because the presence of a word in the subtitle does not guarantee that the word is signed, and variations in signing may occur for the same subtitles.
Second, it is weak because the audio content and the subtitles are not always temporally aligned with the signs in sign language. 
The weak and noisy supervision resulting from these issues ultimately leads to low performance in automatic sign language translation tasks~\cite{Albanie2021bobsl,camgoz2021content4all,varol2021read}.

Interest in automatic sign subtitle annotation has recently increased as a way to effectively utilise data collected from TV broadcasts. Previous works~\cite{varol2021read,albanie2020bsl,momeni2020watch} have primarily concentrated on identifying sparse correspondence between keywords in the subtitles and the signs. To achieve dense alignment between the subtitles and the signs, a method for sign sentence boundary detection based on human body keypoints is proposed in~\cite{bull2020automatic}.
Recently, \cite{bull2021aligning} introduced the task of \textit{aligning subtitles in sign language videos} and developed a baseline model, but no further research has been published on this task due to its inherent difficulty.

In this study, our objective is to develop a framework capable of aligning asynchronous subtitles in sign language videos with limited labelled data. The overall system architecture is visualised in~\Fref{fig:teaser}.
We first introduce a subtitle pre-processing technique that converts natural language sentences into sign language-like sentences, drawing upon fundamental grammatical rules of sign language. It is noteworthy that previous studies in sign language research have overlooked the substantial differences in grammatical structures between sign languages and spoken languages~\cite{sutton1999linguistics}. 
For instance, \cite{bull2021aligning} extract text features using frozen BERT~\cite{devlin2018bert} model pre-trained on a large corpus of natural language data and \cite{varol2021read} employ a straightforward approach of filtering out stop words from subtitles. We reflect the grammar system of sign language, which offers a simpler textual representation compared to natural language, enhancing alignment performance.

We also tackle the challenges posed by the weak and noisy supervision in datasets collected from TV broadcasts. 
To compensate for the noisy supervision caused by ambiguities between sign and subtitle, we introduce a selective alignment loss. This loss function is designed to optimise the model for predicting the precise temporal location of sign language occurrences within the scene, leveraging negative text-video pairs.
By focusing on continuous frames where the sign language matches the queried subtitle, our model can avoid erroneous alignments that do not correspond to the intended text queries.
Additionally, to mitigate the effect of weak supervision stemming from audio-aligned labels, we implement a self-training process. 
Our findings indicate that the accuracy of pseudo-labels generated by our model -- trained on audio-aligned data and fine-tuned on manually labelled data -- is 13.24\% higher than that of audio-aligned labels in frame-level alignment accuracy (75.64\% vs. 62.40\%). As a result, we adopt a self-training strategy where the model is re-trained using the generated pseudo-labels. This iterative process leads to further improvement in model performance.

Our contributions can be summarised as follows: (1) to our best knowledge, we are the first to introduce a subtitle pre-processing algorithm aimed at trimming subtitles by incorporating the grammatical characteristics of sign language, (2) we propose a selective alignment loss, facilitating precise alignment with subtitle by preventing misalignment between unrelated video segment and text query, and (3) we employ a self-training strategy to address the weak supervision arising from heuristic audio-aligned subtitle data.
With extensive experiments, we demonstrate that our proposed method achieves state-of-the-art results in both frame-level accuracy and F1 scores. These findings highlight the effectiveness and practical utility of our framework in advancing the field of sign language video alignment and translation.

\section{Related Works}
\newpara{Sign language spotting.} 
The sign language spotting task involves detecting isolated sign instances within continuous sign language videos.
Early approaches use signing gloves~\cite{liang1998real} and hand-crafted visual elements to capture the hands, face, and motion, integrating these elements with Conditional Random Fields (CRFs)~\cite{yang2006detecting,yang2008sign}, Hidden Markov Models (HMMs)~\cite{Santemiz2009AutomaticSS}, and Hierarchical Sequential Patterns (HSP) Trees~\cite{ong2014sign}. 
Other approaches~\cite{Buehler10, pfister2013large} try to utilise subtitles of broadcast videos as auxiliary supervision to spot signs.
With the advancement of deep learning, more recent work has leveraged extra cues such as mouthings~\cite{albanie2020bsl}
and visual dictionaries~\cite{momeni2020watch} or attached sliding window classifiers~\cite{li2020transferring} to localise signs in time stamps more accurately.
The most recent works~\cite{varol2021read,varol2022scaling}  focus on scaling up sign-spotting with large-scale corpora for automatic sign annotation. However, sign-spotting primarily focuses on detecting individual signs and does not address the broader alignment between video and sentence.

\begin{figure*}[t]
    \centering
    \includegraphics[width=0.9\linewidth]{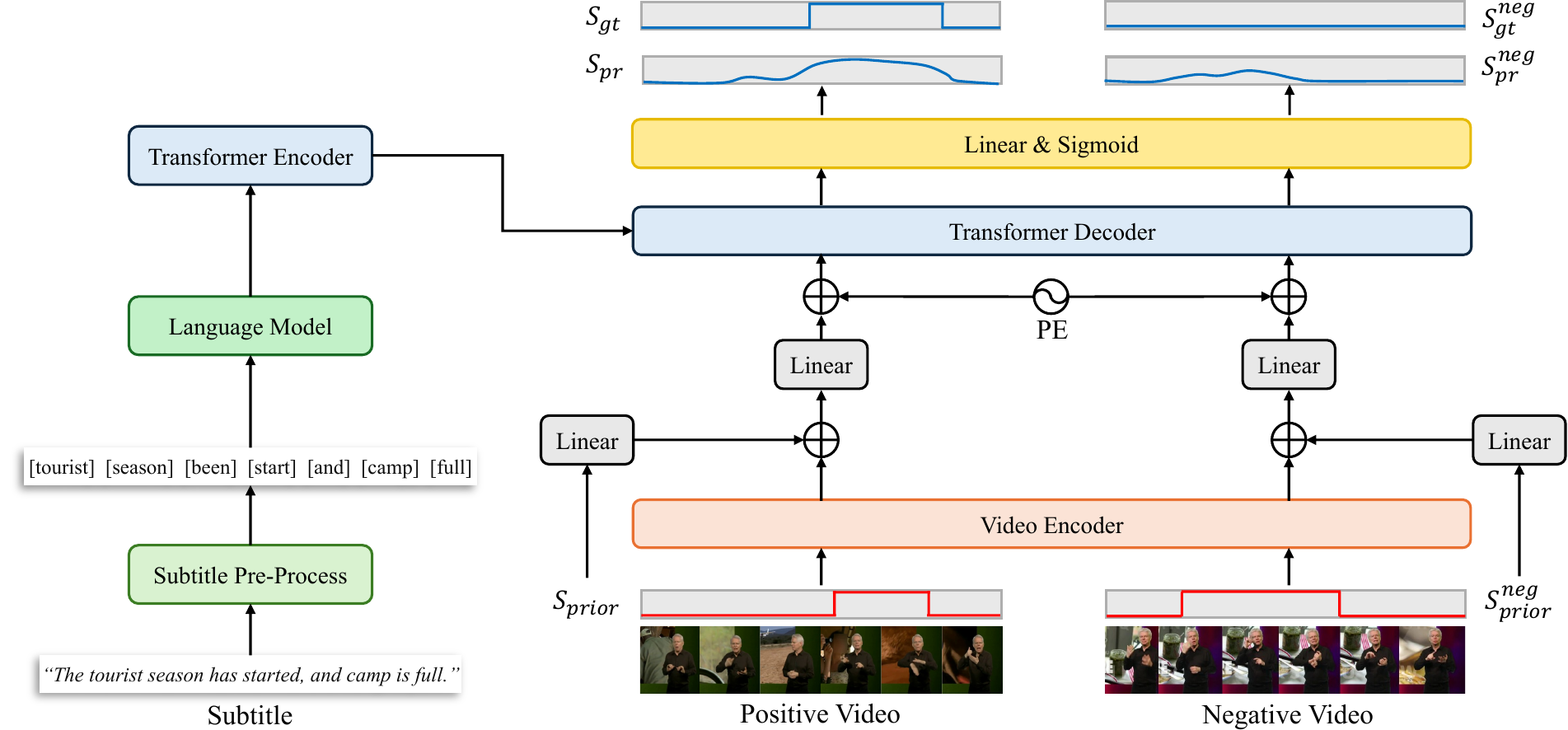}
    \vspace{-2mm}
    \caption{\textbf{An illustration of our framework.} We input to our model: (1) a pseudo-gloss sequence pre-processed by a subtitle pre-processing mechanism, (2) a positive video aligned with the input query, and (3) a negative video not aligned with the input query. Both videos are encoded with shifted temporal boundaries of the audio-aligned subtitle, denoted as $S_{prior}$ and $S^{neg}_{prior}$. Using a Transformer decoder, the model predicts frame-level alignment between text and video. Note that the negative video is only provided during training.}
    \label{fig:overview}
    \vspace{-3mm}
\end{figure*}

\newpara{Continuous sign language recognition.}
The Continuous Sign Language Recognition (CSLR) task aims to map a given sign video to its corresponding gloss\footnote{The smallest unit having independent meaning in sign language.} sequence.
Early works design hybrid models combining CNNs with HMMs ~\cite{koller2016deep,koller2017re}.
Later, Connectionist Temporal Classification (CTC) loss~\cite{graves2006connectionist} is employed to facilitate training of CSLR models~\cite{niu2020stochastic,cui2019deep,jang2022signing}. As a result, it allows the CSLR task to be considered as an alignment task between gloss sequence and sign video. 
From this, one branch focuses on designing model architectures~\cite{min2021visual,ahn2024slowfast,adaloglou2021comprehensive,zhou2021spatial} specialised in temporal alignment, while the other branch~\cite{ koller2016deep,koller2017re,koller2019weakly,jang2023self,hu2023prior} focuses on generating pseudo-gloss labels to propagate frame- or clip-level alignment supervision. 
However, existing methods have primarily been validated on pre-segmented sentences of signing~\cite{camgoz2018neural,ham2021ksl}. This is a major obstacle to applying CSLR technology in the wild, where sign language is continuously streamed.

\newpara{Moment retrieval from text query.}
The Moment Retrieval (MR) task involves identifying temporal moments highly relevant to a given text query within a specified timeframe. This task is typically categorised into two approaches: proposal-based and proposal-free methods. 
Proposal-based methods use a pipeline where candidate windows are generated from the entire video, followed by ranking based on matched scores using predefined temporal structures like sliding windows~\cite{anne2017mcn, gao2017charades, liu2018role, zhang2019tcmn} or temporal anchors~\cite{chen2018tgn, yuan2019scdm, zhang2019man, soldan2021vlg, zhang20202dtan}.
Proposal-free methods treat the task as a regression problem, directly regressing start and end time frames using multimodal attention, dynamic filters, and additional features~\cite{yuan2019ablr, ghosh-etal-2019excl, zhang2020vslnet, mun2020lgi,rodriguez2021dori,chen2021drft,woo2024let}. 
MR shares similarities with sign language subtitle alignment, as both tasks involve aligning text and video at a frame level. However, there are distinct differences:
(1) sign language content requires fine-grained alignment due to the consistent visual appearance of signing sequences across frames, necessitating precise recognition of body dynamics. (2) unlike other alignment tasks, each subtitle in sign language alignment has a specific reference location, which provides prior information about its start time and duration. In this paper, our primary focus is on the sign language subtitle alignment task, addressing the unique challenges and considerations inherent to this domain.

\newpara{Sign language subtitle alignment.}
Early work~\cite{farhadi2006aligning} on aligning subtitles to continuous signing combined cues from multiple sparse correspondences. However, this work was based on the assumption that the ordering of words in subtitles is the same as that of signing. Since then, the sequence-level sign language temporal localisation task has been studied using various approaches such as category-agnostic sign segmentation~\cite{farag2019learning,renz2021sign}, signer diarisation~\cite{gebrekidan2013automatic,gebre2014motion,albanie2021signer}, and active signer detection~\cite{cherniavsky2008activity,borg2019sign,moryossef2020real,shipman2017speed}. 
Nevertheless, these studies deal with temporal granularity, such as word boundaries or active sign segments, rather than subtitle units.
In response to this, \cite{bull2020automatic} initially introduce a model for segmenting sign language video into sentence-like units. While the model is accurate at detecting temporal boundaries, achieving an ROC-AUC metric of 0.87 at the frame level, aligning these segmented units with corresponding subtitles requires additional steps such as sign language translation~\cite{camgoz2018neural,li2020tspnet,camgoz2021content4all,chen2022simple,jang2025lost,zhou2021spatial} and text processing.
More recently, the sentence-level sign alignment task has been extended by simulating real-world data acquisition scenario~\cite{bull2021aligning} using a large-scale broadcast dataset~\cite{Albanie2021bobsl}. However, other than this study, research on sign language subtitle alignment in real-life scenarios defined in~\cite{bull2021aligning} has not been actively conducted due to the high entry barrier posed by the difficulty of this task.
In this paper, we present a novel task-specific framework that significantly outperforms the baseline's alignment performance with a linguistic analysis of British Sign Language.
\section{Method}

\subsection{Sign to subtitle alignment framework} \label{sec:framework}
Our framework is illustrated in~\Fref{fig:overview}. For the input of the model, we provide (1) a pseudo-gloss sequence pre-processed by a subtitle pre-processing algorithm, (2) a positive video aligned with the input query, and (3) a negative video not aligned with the input query. Since guiding the model with prior which is a temporal boundary from the audio-aligned subtitle has a significant impact on model performance~\cite{bull2021aligning}, both videos are encoded with the priors denoted as $S_{prior}$ and $S^{neg}_{prior}$. Finally, the model produces a vector of values ranging from 0 to 1, indicating the relevance of each frame to the text query. The first and last values exceeding a threshold $\tau$ define the predicted temporal boundaries for the query subtitle. Notably, during inference, only the positive video is inputted.

\newpara{Subtitle pre-processing.} 
With an understanding of British Sign Language (BSL) linguistics, we transform the text query input into a sign language-like sentence. Further details about this process will be provided in~\Sref{sec:preprocess}.

\newpara{Text encoding.} 
The pre-processed subtitle is encoded with a language model pre-trained on a large text corpus. This model generates a sequence of features, with special tokens representing the beginning and end of a sentence. To align with the input dimension of the Transformer encoder, these sequence features are projected to $d_{model}$ using a linear layer. Subsequently, the output from the linear layer is fed into the Transformer~\cite{vaswani2017attention} encoder to propagate token-level correlation information for the decoding process.

\newpara{Video encoder.} 
We input a video sequence of length $T$ into the visual encoder for feature extraction. The encoded features are projected to $d_{model}$ to be combined with the text modality in the Transformer decoder.

\newpara{Prior encoding.} 
Following the approach described in~\cite{bull2021aligning}, we input audio-aligned subtitle timing as a prior position and duration cue. The prior estimate is encoded as a binary vector of length $T$, where 1 indicates that the video frame is within the subtitle's temporal boundaries, and 0 otherwise. This prior estimate is then projected to a dimension $d_{model}$, concatenated with the video features along the channel axis, and finally projected again to $d_{model}$.

\newpara{Multimodal decoding.} 
The decoder comprises Transformer layers that process the encoded sequence to aggregate visual and text features. Positional encoding is applied to capture the temporal order of signing.
The final layer is a linear transformation followed by a sigmoid activation, producing $T$ predictions between $[0, 1]$ for each video frame. 

\subsection{Subtitle pre-processing} \label{sec:preprocess}
The grammatical system of  British Sign Language (BSL) significantly differs from that of spoken languages~\cite{sutton1999linguistics}. However, existing deep learning-based sign language-related studies simply adopt techniques from the Natural Language Processing (NLP) field, such as freezing pre-trained language models or reducing noise in the input text. 
In this study, we explore the impact of sign language grammar rules on alignment performance and propose an online subtitle pre-processing algorithm. This algorithm transforms natural language sentences into sentences that mimic the grammar system of sign language, in other words, resemble gloss representation. 
We delve into the grammar of BSL based on insights from~\cite{sutton1999linguistics}. 
We define guidelines that characterise the unique linguistic structure of BSL: (1) pronouns extend beyond referring to people and also include concepts of places where people exist, (2) negative meaning is conveyed through negative phrases, (3) articles may be omitted in gloss representation, (4) tense may not always be explicitly indicated, (5) {\color{magenta}`be'} verbs may be omitted in gloss representation, and (6) {\color{blue}`been'} word is used to express the present perfect tense. 
Reflecting on these rules,  the following strategies are adopted: (1-2) we do not remove stopwords that are considered an unimportant component in NLP field, (3) exclude articles, (4) lemmatise verbs (5) eliminate {\color{magenta}`be'} verbs, and (6) replace the auxiliary verb {\ colour {orange}`have'} with {\color{blue}`been'}. 
At the last stage, we analyse the role of the word {\color{orange}`have'} in a sentence using the Part of Speech (PoS) tagging approach. 
Our experiments prove that tailoring input queries based on the grammatical rules of BSL significantly enhances alignment performance. This result underscores the importance of incorporating sign language-specific linguistic features in the subtitle alignment domain.

\subsection{Training and fine-tuning} \label{sec:training}
The sign language subtitle alignment task ultimately aims to achieve auto-labelling while minimising the reliance on extensive human labelling. In this study, we simulate a scenario where the amount of manually labelled data is significantly less compared to the volume of audio-aligned labelled data. This approach reflects real-world challenges where acquiring large-scale manually labelled datasets for sign language alignment is resource-intensive and impractical.

Our training strategies are composed of three stages.
Firstly, we perform word pre-training, a task involving single sign-spotting under the same setup of~\cite{bull2021aligning}. 
Notably, our model can be applied to the sign-spotting task without any modification, except for the input data type (sentence $\rightarrow$ word), achieving improved performance compared to the previous method. Experimental results are detailed in~\Sref{sec:spotting}. 
Secondly, we conduct a training stage using weak labels, specifically audio-aligned subtitles. This stage focuses on helping the model learn from various text corpus. 
Finally, to refine model predictions, we fine-tune the model using a small amount of manually labelled subtitle data. During these phases, the model is trained with a binary cross-entropy loss between the predicted vector $S_{pr}$ and the ground truth $S_{gt}$, defined as follows:
\begin{equation}
    \small
    \mathcal{L}_{align}=- \frac{1}{T} \sum_{t=1}^{T} \left(
    S_{gt,t} \log S_{pr,t} + (1 - S_{gt,t}) \log (1 - S_{pr,t})
    \right),
\end{equation}
where $t$ denotes a frame index. This loss encourages our model to discriminate which frames are relevant to the subtitle and which are not.

\subsection{Selective alignment loss} \label{sec:loss}

As explained in~\Sref{sec:training}, our model is trained on audio-aligned subtitles in the training stage. 
In practice, $S_{prior}$ is the timing of the temporally jittered audio-aligned subtitle. 
However, this setup leads to increased dependence on prior information. 
To solve this problem, this paper introduces selective alignment loss, which is composed of negative alignment loss and relative alignment loss.

\newpara{Negative alignment loss.}
To reduce dependence on the prior, we introduce negative video-subtitle pairs to simulate scenarios where our model should predict non-aligned frames based on text understanding. 
In our implementation, when the model receives a negative pair as input, the ideal output should be a zero-valued sequence vector because there is no video frame associated with the given subtitle. 
We use binary cross-entropy loss to train the model with a zero-valued sequence vector as the label for negative video pairs. 
The negative alignment loss can be formulated as follows:
\begin{equation}
    \mathcal{L}_{neg}=- \frac{1}{T} \sum_{t=1}^{T}
    \log(1 - S^{neg}_{pr,t}),
\end{equation}
where $S^{neg}_{pr}$ represents the model prediction when the negative pair is given as the input. This loss function helps the model learn to distinguish misaligned video-subtitle pairs more effectively, thereby improving its alignment capabilities.

\newpara{Relative alignment loss.}
When negative samples are given into the model during training, a bias arises where the model tends to output more zero-valued sequences. To address this issue, inspired by~\cite{khosla2020supervised}, we propose a relative alignment loss that leverages the provided ground truth labels to differentiate between probabilities of aligned frames from those of non-aligned frames.
To compute the relative importance, the proposed loss is defined as follows:
\begin{equation}
    \mathcal{L}_{rel}=- \frac{1}{\sum_{t=1}^{T}S_{gt,t}} \sum_{t=1}^TS_{gt,t} \text{log}\frac{
    \mathrm{e}^{S_{pr,t}}}
    {
    \sum_{i=1}^{T}\mathrm{e}^{S_{pr,i}} + \mathrm{e}^{S^{neg}_{pr,i}}
    }.
\end{equation}

This loss function aims to mitigate the aforementioned challenges and encourage the model to predict alignments more effectively, reducing sensitivity to the number of aligned frames in the ground truth. The total loss $\mathcal{L}_{tot}$ is as follows:
\begin{equation}
    \mathcal{L}_{tot} = \mathcal{L}_{align} + \lambda_{neg} \mathcal{L}_{neg} 
    + \lambda_{rel} \mathcal{L}_{rel},
\end{equation}
where $\lambda_{neg}$ and $\lambda_{rel}$ are hyperparameters for negative alignment loss and relative alignment loss respectively. In our experiments, $\lambda_{neg}$ and $\lambda_{rel}$, are set to 1.

\subsection{Self-training} \label{sec:self-training}
In this study, our model is trained using an audio-aligned subtitle temporal boundary that can be obtained relatively easily and fine-tuned using a small amount of human-labeled data. 
Although we achieve improved alignment performance through fine-tuning, we once again consider the negative impact of weak labels used in the training stage. 
To address this issue, we focus on the self-training technique, which involves labelling unlabeled data using a fully trained model and subsequently retraining the model with this data.
Our self-training technique in the alignment problem comprises of two steps. Firstly, we derive frame-level alignment confidence from sign language video clips and text query pairs. Subsequently, we compare the peak confidence score against a predefined confidence threshold $\tau_{c}$ to determine whether to utilise it as self-training data. In essence, data with a peak confidence score below the threshold is filtered out.
Furthermore, we analyse the correlation between data quality and quantity by adjusting the threshold $\tau_{c}$. Our experiments demonstrate that self-training effectively enhances model performance.

\section{Experiments}

\subsection{Experimental setup} \label{sec:experimental_setup}
\newpara{Dataset.}
Our framework is trained on the BOBSL dataset~\cite{Albanie2021bobsl}, a publicly available dataset comprising British Sign Language interpreted BBC broadcast footage, accompanied by English subtitles corresponding to the audio content. This dataset consists of 1,940 episodes totaling 1,447 hours, with 1.2M sentences covering a vocabulary of 77K words. It involves 37 signers. 
The experimental setup, including the dataset split, follows the publicly available code of~\cite{bull2021aligning} provided by their official repository\footnote{\url{https://github.com/hannahbull/subtitle_align}}.
We use a subset of this dataset, specifically 16 episodes, for fine-tuning our model and 35 episodes for testing. The test set encompasses 30 hours of video and includes 20,338 English subtitles, with a total vocabulary of 13K words.

In this research, we hypothesise that the cost of manually labelling alignment data can be minimised by leveraging a small amount of human-labelled data, representing approximately 2\% of the volume of heuristic audio-aligned data available. This approach demonstrates our strategy to efficiently utilise limited labelled data to achieve state-of-the-art performance in sign language alignment tasks.

\newpara{Evaluation protocol.}
We follow the evaluation protocol proposed in~\cite{bull2021aligning}, which utilise frame-level accuracy and F1-score to assess the performance of alignment models. For evaluating the F1-score, we quantify the accuracy of subtitle alignment with sign language videos by considering hits and misses based on temporal overlaps. Specifically, we use three temporal overlap thresholds (IoU $\in~\{0.1, 0.25, 0.50\}$, representing F1@.10, F1@.25, and F1@.50, respectively). These metrics compare the predicted spans S$_{pr}$ with ground truth subtitle spans $S_{gt}$ to measure alignment accuracy. The frame-level accuracy is denoted as frame-acc in our tables.
For evaluating the sign-spotting task, we calculate two metrics: mean average precision (mAP) and top-1 classification accuracy (Acc@1).

\newpara{Implementation details.} 
In subtitle pre-processing, we utilise the pycontractions library\footnote{\url{
https://pypi.org/project/pycontractions}} for fixing contractions and the Natural Language Toolkit~\cite{bird2009natural} for lemmatisation and Part of Speech (PoS) tagging.
For model architecture, we follow the architecture of SAT model.
The language model is based on the BERT architecture, pre-trained on BookCorpus and English Wikipedia, featuring 12 layers, 12 attention heads, and a model size of 768.
We use a vanilla 2-layer Transformer encoder and decoder with 4 attention heads, where the Transformer's model dimension $d_{model}$ is set to 512.
The video encoder leverages a pre-trained I3D~\cite{carreira2017quo} sign classification model~\cite{varol2021read}, producing 1024-dimensional visual embeddings pre-extracted from sign language video segments.
For positional encodings in the input to the video encoder, we employ 512-dimensional sinusoidal positional encodings.
The prior subtitle temporal boundary $S_{prior}$ is derived from the temporal location of the audio-aligned subtitle $S_{audio}$ shifted by $+2.7$ seconds then spanned $3.2$ seconds on each side, denoted as $S^{+}_{audio}$. 
During training, we randomly select a $20$-second search window around the location of the ground truth subtitle $S_{gt}$. Within this window, we sample video features with a stride of $4$. Note that $S_{gt}$ represents an audio-aligned subtitle during training and transitions to a sign-aligned subtitle during fine-tuning. 
To ensure a fair comparison with previous methods, we use a DTW threshold of 0.4 for global alignment in long-sequence videos.
We use the Adam~\cite{kingma2014adam} optimiser with a batch size of 64 for training. During the word pre-training stage, the learning rate is set to $10^{-5}$, and for training and fine-tuning with subtitles, we use a learning rate of $5\times10^{-6}$. The word pre-training stage involves training the model for 7 epochs, followed by 4 epochs of training for subsequent stages. Full-sentence fine-tuning is conducted over 100 epochs.

\begin{table}
  \renewcommand{\arraystretch}{1.2}
  \renewcommand{\tabcolsep}{2mm}
  \centering
  \vspace{-1mm}
  \caption{\textbf{Performance comparison to baselines.} Our method outperforms existing approaches across all metrics, demonstrating its capability to align subtitles with sign videos.}  
  \resizebox{0.9\linewidth}{!}{
  \begin{tabular}{lcccc}
    \toprule
    Method & frame-acc ($\%$) $\uparrow$ & F1@.10 $\uparrow$ & F1@.25 $\uparrow$ & F1@.50 $\uparrow$ \\
    \midrule
    $S_{audio}$ & 40.14 & 46.36 & 33.46 & 14.10 \\
    $S^+_{audio}$ & 62.40 & 72.79 & 64.09 & 44.60 \\
    SAT \cite{bull2021aligning} & 71.01 & 74.19 & 67.01 & 53.36 \\
    \midrule
    \textbf{Ours} & \textbf{77.22} & \textbf{81.39} & \textbf{75.03} & \textbf{63.81} \\
    \bottomrule
  \end{tabular}}
  \label{tab:main}
  \vspace{-1mm}
\end{table}
\begin{table}[t]
  \renewcommand{\arraystretch}{1.2}
  \renewcommand{\tabcolsep}{2mm}
  \centering
  \caption{\textbf{Evaluation on downstream tasks.} The pseudo-subtitles generated by our framework contribute to performance improvement in both retrieval and continuous sign language recognition (CSLR) tasks.}
  \vspace{-1mm}
  \resizebox{0.9\linewidth}{!}{
  \begin{tabular}{l|cccc|c}
    \toprule
    \multicolumn{1}{l|}{Task} & \multicolumn{4}{c|}{Retrieval} & \multicolumn{1}{c}{CSLR} \\
    \midrule
    \multirow{1}{*}{Metric} & \multicolumn{2}{c}{T2V} & \multicolumn{2}{c|}{V2T} & \multicolumn{1}{c}{WER}$\downarrow$ \\
    & R@1$\uparrow$ & R@5$\uparrow$ & R@1$\uparrow$ & R@5$\uparrow$ & \\
    \midrule
    $CSLR^{2}$~\cite{raude2024} w/SAT~\cite{bull2021aligning} & 27.14 & 42.19 & 26.25 & 41.99 & 65.16 \\
    $CSLR^{2}$~\cite{raude2024} w/Ours & \textbf{28.59} & \textbf{43.74} & \textbf{26.59} & \textbf{42.51} & \textbf{64.22} \\
    \bottomrule
  \end{tabular}}
  \label{tab:cslr}
\end{table}
\subsection{Quantitative results} \label{sec:quantitative}
\newpara{Comparison to baselines.}
As shown in~\Tref{tab:main}, first, we evaluate two baseline approaches: using the original, non-shifted audio-aligned subtitle $S_{audio}$ and using the shifted audio-aligned subtitle $S^{+}_{audio}$. The results demonstrate significant performance improvements by simply shifting the time of the audio-aligned subtitle.
Next, we assess the performance of the SAT~\cite{bull2021aligning} model, which represents the initial work in this field. The SAT model outperforms all baselines by leveraging subtitle text to identify associated video segments. Finally, our proposed method exhibits superior performance across all metrics by a substantial margin. This underscores our framework's capability to align subtitles with sign videos based on a deep understanding of BSL's linguistic characteristics. Note that all metric is calculated with Dynamic Time Warping (DTW) for global alignment in long-time sequence videos with overlapped regions removed, following the method described in~\cite{bull2021aligning}.



\begin{table}[t]
  \renewcommand{\arraystretch}{1.2}
  \renewcommand{\tabcolsep}{2.5mm}
  \centering
  \vspace{-1mm}
  \caption{\textbf{Sign word spotting performance.} We randomly jitter prior $S_{prior}$. Our model shows superior performance in the sign spotting task.}
  \resizebox{0.6\linewidth}{!}{
  \begin{tabular}{lccc}
    \toprule
    Method & mAP $\uparrow$ & Acc@1 ($\%$) $\uparrow$ \\
    \midrule
    Random prior & 7.60 & 4.67 \\
    SAT \cite{bull2021aligning} & 70.83 & 68.94 \\
    \textbf{Ours} & \textbf{76.89} & \textbf{75.35} \\
    \bottomrule
  \end{tabular}}
  \label{tab:spotting}
  \vspace{-1mm}
\end{table}
\newpara{Evaluation on downstream tasks, sign language retrieval and continuous sign language recognition.}
For the baseline of both retrieval and Continuous Sign Language Recognition (CSLR) tasks, we utilize the $CSLR^{2}$ model~\cite{raude2024}, which is trained using pseudo-subtitle labels generated by the SAT model. We train the $CSLR^{2}$ model from scratch with automatically generated subtitles produced by our pipeline.
For retrieval evaluation, we report both text-to-video (T2V) and video-to-text (V2T) performance using standard retrieval metrics, specifically \textbf{recall at rank k (R@k)}, where k $\in$ {1,5}. For CSLR evaluation, we report the word error rate (WER).

As shown in~\Tref{tab:cslr}, the improved subtitle quality leads to better performance across all metrics. Based on this, we believe that our method can be applied to various sign language downstream tasks by providing more reliable pseudo-subtitles for model training.

\newpara{Sign-spotting performance.}
As mentioned earlier, our model can be directly applied to the sign-spotting task without changing the model architecture. To use the model in this way, we input a random prior into the model during both the training and testing stages. As shown in~\Tref{tab:spotting}, the accuracy of the random prior is very low at 7.6 based on mAP. In contrast, the SAT model achieves an mAP of 70.83 and Acc@1 of 68.94. Our model achieves even higher performance with an mAP of 76.89 and Acc@1 of 75.35. This result demonstrates the general applicability of our model to tasks requiring an understanding of sign language.
\label{sec:spotting}

\newpara{Comparison with gloss representation.}
We train the model using glosses instead of subtitles. As there is no publicly available text-to-gloss model for BSL, we utilise pseudo-glosses extracted by~\cite{momeni2022automatic}.
As shown in~\Tref{tab:gloss}, our approach shows better performance, enabling the model to understand sign language grammar with supervision. Note that we do not conduct self-training on both models.
\begin{table}
  \renewcommand{\arraystretch}{1}
  \renewcommand{\tabcolsep}{1.1mm}
  \centering
  \vspace{-1mm}
  \caption{\textbf{Comparison with gloss.} We emphasise that our subtitle pre-processing does not require any training stage or manual labelling process, unlike gloss production.}
  \resizebox{0.90\linewidth}{!}{
  \begin{tabular}{l|cccc}
    \toprule
    Pre-processing & frame-acc ($\%$) $\uparrow$ & F1@.10 $\uparrow$ & F1@.25 $\uparrow$ & F1@.50 $\uparrow$ \\
    \midrule
    Gloss & 72.92 & 78.43 & 71.10 & 57.12 \\
    Subtitle (\bf{Ours}) & \textbf{75.64} & \textbf{80.02} & \textbf{73.35} & \textbf{61.45} \\
    \bottomrule
  \end{tabular}}
  \label{tab:gloss}
\end{table}

\newpara{Analysis of input data at inference.}
We create subsets from the BOBSL test set through various filtering strategies and explore the corresponding performance variations. When evaluating only the data overlapping with $S_{prior}$ (Overlap only case), there is a significant performance boost of 12.35 in F1@0.50. This demonstrates that a better prior can contribute to improved performance and highlights the importance of properly setting the prior in the training data. Furthermore, when removing cases with insufficient language information—specifically subtitles with fewer words (Remove shorter (n) cases)—we observe an overall improvement in performance. This proves that a lack of language information negatively impacts model performance. Lastly, when removing cases with longer subtitles (Remove longer (n) cases), performance slightly decreases, indicating that an excessive amount of language information does not necessarily contribute to further improvements.

\begin{table}[t]
  \renewcommand{\arraystretch}{1.3}
  \renewcommand{\tabcolsep}{1.2mm}
  \centering
  \vspace{-1mm}
  \caption{\textbf{Analysis of input data at inference.} \textit{All}: the entire dataset; \textit{Overlap only}: cases where there is an overlap between $S_{prior}$ and the subtitle; \textit{Remove shorter ($n$)}: excludes subtitles with fewer than $n$ words; \textit{Remove longer ($n$)}: removes subtitles with more than $n$ words.}
  \resizebox{0.99\linewidth}{!}{
  \begin{tabular}{l|c|cccc}
    \toprule
    Data & \#sent & frame-acc ($\%$) $\uparrow$ & F1@.10 $\uparrow$ & F1@.25 $\uparrow$ & F1@.50 $\uparrow$ \\
    \midrule
    All & 20,338 & 77.22 & 81.39 & 75.03 & 63.81 \\
    Overlap only & 15,898 & 80.80 & 91.27 & 86.81 & 76.16 \\
    Remove shorter (2) & 19,204 & 77.55 & 83.71 & 77.95 & 66.64 \\
    Remove shorter (3) & 17,817 & 77.95 & 85.44 & 80.25 & 68.96 \\
    Remove shorter (4) & 16,406 & 78.48 & 87.02 & 82.44 & 71.22 \\
    Remove longer (50) & 20,280 & 77.35 & 81.30 & 74.89 & 63.59 \\
    \bottomrule
  \end{tabular}}
  \label{tab:component}
  \vspace{-1mm}
\end{table}

\begin{table}[t]
  \renewcommand{\arraystretch}{1.3}
  \renewcommand{\tabcolsep}{1.2mm}
  \centering
  \vspace{-1mm}
  \caption{\textbf{Ablation study on model component.} we demonstrate that each element -- subtitle pre-processing, selective alignment loss, and self-training -- consistently contributes to performance improvements.}
  \resizebox{0.999\linewidth}{!}{
  \begin{tabular}{l|cccc}
    \toprule
    Method & frame-acc ($\%$) $\uparrow$ & F1@.10 $\uparrow$ & F1@.25 $\uparrow$ & F1@.50 $\uparrow$ \\
    \midrule
    Baseline & 72.59 & 77.04 & 69.84 & 56.24 \\
    + subtitle pre-processing & 74.46{\small\,\textcolor{blue}{(+1.87)}} & 78.62{\small\,\textcolor{blue}{(+1.58)}} & 71.75{\small\,\textcolor{blue}{(+1.91)}} & 59.50{\small\,\textcolor{blue}{(+3.26)}} \\
    \,\,\,+ selective alignment loss & 75.64{\small\,\textcolor{blue}{(+3.05)}} & 80.02{\small\,\textcolor{blue}{(+2.98)}} & 73.35{\small\,\textcolor{blue}{(+3.51)}} & 61.45{\small\,\textcolor{blue}{(+5.21)}} \\
    \,\,\,\,\,\,+ self-training \textbf{(Ours)} & \textbf{77.22}{\small\,\textcolor{blue}{(+4.63)}} & \textbf{81.39}{\small\,\textcolor{blue}{(+4.35)}} & \textbf{75.03}{\small\,\textcolor{blue}{(+5.19)}} & \textbf{63.81}{\small\,\textcolor{blue}{(+7.57)}} \\
    \bottomrule
  \end{tabular}}
  \label{tab:component}
\end{table}

\subsection{Ablation study} \label{sec:ablation}
\newpara{Effectiveness of model component.} 
We verify the effectiveness of the three components comprising our framework—subtitle pre-processing, selective alignment loss, and self-training—through various ablation studies. As reported in~\Tref{tab:component}, both frame-level accuracy and F1 scores demonstrate improvement when utilising subtitle pre-processing that incorporates the linguistic characteristics of sign language. Furthermore, additional performance gains are observed with the inclusion of selective alignment loss, which facilitates learning the correlation between text and video using negative pairs.
Finally, alignment performance is notably enhanced through a self-training process, where our model is trained with its own outputs instead of relying solely on weakly labelled audio-aligned subtitles. This analysis demonstrates that each component comprising our model contributes to improving overall performance.

\begin{table}[t!]
  \renewcommand{\arraystretch}{1.2}
  \renewcommand{\tabcolsep}{1.2mm}
  \centering
  \vspace{-1mm}
  \caption{\textbf{Ablation study on effectiveness of each process in subtitle pre-processing.} This shows incremental performance improvements with each added process, achieving the best performance when all processes are applied.}
  \resizebox{0.99\linewidth}{!}{
  \begin{tabular}{l|cccc}
    \toprule
    Method & frame-acc ($\%$) $\uparrow$ & F1@.10 $\uparrow$ & F1@.25 $\uparrow$ & F1@.50 $\uparrow$ \\
    \midrule
    Baseline with SA loss & 73.77 & 76.93 & 69.95 & 57.36 \\
    + stopwords & 75.23 & 79.57 & 73.02 & 60.72 \\
    ~+ remove articles & 75.43 & 79.72 & 73.45 & 60.85 \\
    ~~+ lemmatise & 75.46 & 79.88 & 73.48 & 61.21 \\
    ~~~+ `be' verb process \textbf{(Ours)} & 75.64 & 80.02 & 73.35 & 61.45 \\
    \bottomrule
  \end{tabular}}
  \label{tab:textpreprocess}
  \vspace{-2mm}
\end{table}
\newpara{Effectiveness of subtitle pre-processing.}
In~\Tref{tab:textpreprocess}, we investigate the impact of reflecting sign language grammar in subtitle pre-processing on the model's understanding of sign language.
Instead of removing stopwords (including fixed contractions), we observe a notable performance improvement of 1.46\% compared to the baseline. This finding suggests that stopwords carry significant meaning in sign language, unlike in traditional NLP tasks. The removal of articles and lemmatisation of verbs also contribute to a steady improvement in performance, indicating the importance of these pre-processing steps in enhancing the model's comprehension of signed language. The process of removing the `be' verb and incorporating `been' proves beneficial for the model's understanding of sign language. This adjustment improves the model's ability to capture sign language grammar and nuances effectively.
Overall, these findings highlight the importance of tailored subtitle pre-processing techniques that align with the grammar and structure of signed language. Optimising these pre-processing steps can significantly enhance the model's performance and understanding of sign language.

\begin{table}[t]
  \renewcommand{\arraystretch}{1.2}
  \centering
  \vspace{-1mm}
  \caption{\textbf{Ablation results according to the fine-tuning strategies.} Our final model freezes both Language Model and Transformer Encoder during the fine-tuning stage to prevent overfitting to a limited amount of labelled data.}
  \resizebox{0.999\linewidth}{!}{
  \begin{tabular}{cc|cccc}
    \toprule
    \multicolumn{2}{c|}{\textbf{Freeze}} & \multirow{2.65}{*}{frame-acc (\%) $\uparrow$} & \multirow{2.65}{*}{F1@.10 $\uparrow$} & \multirow{2.65}{*}{F1@.25 $\uparrow$} & \multirow{2.65}{*}{F1@.50 $\uparrow$} \\
    \makecell{Language\\Model} & \makecell{Transformer\\Encoder} &&&& \\
    \midrule
    & & 75.04 & 79.01 & 72.59 & 60.58 \\
    \cmark & & \textbf{75.70} & 79.93 & 73.26 & 61.37 \\
    \cmark & \cmark & 75.64 & \textbf{80.02} & \textbf{73.35} & \textbf{61.45} \\
    \bottomrule
  \end{tabular}}
  \label{tab:text_model}
\end{table}

\newpara{Fine-tuning of subtitle encoder.} 
In this paper, we explore the grammatical and linguistic system of BSL. This results in a unique form of text input that significantly differs from natural language. Therefore, we highlight the necessity of going beyond simply using conventional models used in the NLP field during the text encoding process. Instead, we emphasise the importance of developing a sign language expert model, tailored to handle the intricacies of sign language linguistics.

We conduct ablation studies by freezing the language model, which is a BERT~\cite{devlin2018bert} model pre-trained on various vocabularies, and the Transformer encoder, which maps the output of the language model to features useful for the alignment task during our fine-tuning stage. As shown in~\Tref{tab:text_model}, training both the language model and Transformer encoder during fine-tuning results in relatively low performance. We expect that the model is overfitted on a limited amount of labelled data. Additionally, we observe that freezing both the language model and transformer encoder during fine-tuning yields similar performance to freezing only the language model.
This result demonstrates the effectiveness of the strategy to learn a diverse vocabulary during the training stage of subtitle alignment. By training the model to retain this learned vocabulary and avoid forgetting it during fine-tuning with a small amount of data, we achieve improved performance.

\begin{table}
  \renewcommand{\arraystretch}{1.2}
  \centering
  \vspace{-1mm}
  \caption{\textbf{Ablation results according to the amount of the data used in the self-training stage.} The model's performance steadily improves with more data, which indicates that the output of our model is reliable.}
  \resizebox{0.999\linewidth}{!}{
  \begin{tabular}{cc|cccc}
    \toprule
    \multirow{2}{*}{\makecell{Confidence\\threshold}} & \multirow{2}{*}{Data ratio} & \multirow{2}{*}{frame-acc (\%) $\uparrow$} & \multirow{2}{*}{F1@.10 $\uparrow$} & \multirow{2}{*}{F1@.25 $\uparrow$} & \multirow{2}{*}{F1@.50 $\uparrow$} \\
    &&&&&\\
    \midrule
    \rowcolor{lightgray!60} \multicolumn{2}{c|}{Train with $S^{+}_{audio}$} & 75.50 & 79.73 & 73.29 & 61.29 \\
    \midrule
    0 & 100.0\% & \textbf{77.22} & \textbf{81.39} & \textbf{75.03} & \textbf{63.81} \\
    0.5 & 88.0\% & 77.09 & 80.91 & 74.70 & 63.51 \\
    0.9 & 62.2\% & 76.94 & 80.76 & 74.54 & 63.46 \\
    0.95 & 49.2\% & 76.60 & 80.11 & 73.96 & 62.78 \\
    \bottomrule
  \end{tabular}}
  \label{tab:self_training}
  \vspace{-2mm}
\end{table}

\begin{table}
  \renewcommand{\arraystretch}{1.2}
  \renewcommand{\tabcolsep}{2mm}
  \centering
  \vspace{-1mm}
  \caption{\textbf{Ablation study on selective alignment loss.} The best performance is achieved when both negative alignment loss $\mathcal{L}_{neg}$ and relative alignment loss $\mathcal{L}_{rel}$ are applied.}
  \resizebox{0.9\linewidth}{!}{
  \begin{tabular}{cc|cccc}
    \toprule
    $\mathcal{L}_{neg}$ & $\mathcal{L}_{rel}$ & frame-acc ($\%$) $\uparrow$ & F1@.10 $\uparrow$ & F1@.25 $\uparrow$ & F1@.50 $\uparrow$ \\
    \midrule
     & & 74.46 & 78.62 & 71.75 & 59.50 \\
     & \cmark & 74.32 & 78.34 & 71.75 & 59.36  \\
    \cmark & & 74.94 & 78.23 & 71.49 & 59.19 \\
    \cmark & \cmark & \textbf{75.64} & \textbf{80.02} & \textbf{73.35} & \textbf{61.45}\\
    \bottomrule
  \end{tabular}}
  \label{tab:ablation_neg_bal}
  \vspace{-2mm}
\end{table}

\newpara{Ablation study on the amount of data used in self-training.}
We demonstrate the reliability of the fine-tuned model's outputs used in self-training, as detailed in~\Tref{tab:self_training}. 
Our methodology involves feeding text queries from audio-aligned data alongside corresponding videos into the model, followed by an assessment of peak confidence levels. Only data exceeding the confidence threshold are integrated into the self-training process.
Re-training the model with heuristic audio-aligned labels (denoted as `Train with $S^{+}_{\text{audio}}$') under the same setting with self-training does not result in performance improvement (refer to the performance of `Baseline + subtitle pre-processing + selective alignment loss' in~\Tref{tab:component}). 
Through experiments involving the adjustment of the confidence threshold, we establish that greater performance gains are achieved in the self-training process with an increased volume of pseudo-labels.
This finding underscores that even when the model's alignment confidence is low, the quality of pseudo-labels from our model surpasses that of heuristic audio-aligned subtitles.

\begin{figure*}[t]
    \centering
    \includegraphics[width=0.95\linewidth]{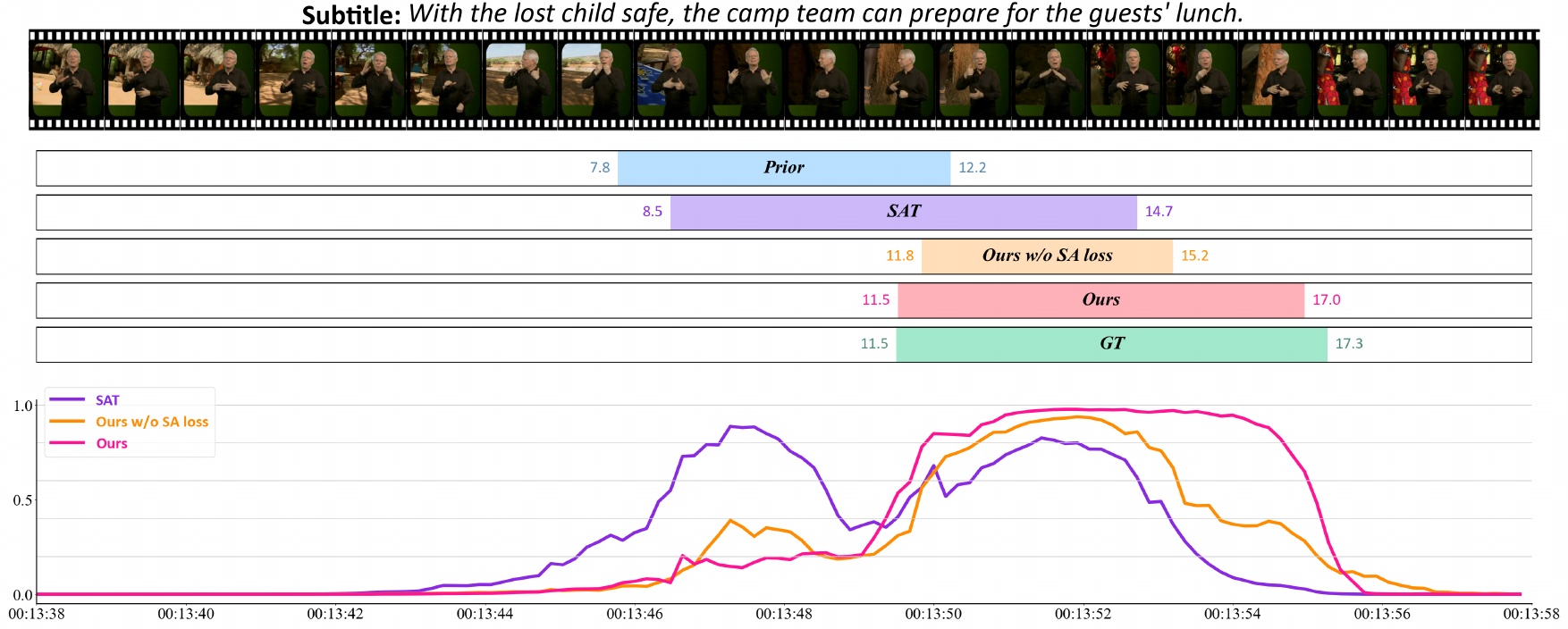}
    \vspace{-1mm}
    \caption{Qualitative results. This figure presents frame-level model predictions, showcasing the timeline of prior, the SAT model, our model trained without the selective alignment loss, our full model, and ground truth. Our model, which incorporates rich sign language linguistics, shows superior alignment with the ground truth compared to other baselines. 
    In addition, the frame-level alignment probability visualisation at the bottom of this figure illustrates our model's clear timeline alignment, demonstrating its effective correlation of text with video content.}
    \label{fig:qualitative}
  \vspace{-2mm}
\end{figure*}

\newpara{Ablation study on selective alignment loss.}
In~\Tref{tab:ablation_neg_bal}, we analyse the effects of negative alignment loss $\mathcal{L}_{neg}$ and relative alignment loss $\mathcal{L}_{rel}$ that constitute the proposed selective alignment loss. When using only $\mathcal{L}_{rel}$, we observe a slight degradation in model performance. This decrease is expected because it unnecessarily restricts the model's ability to predict a balanced ratio of $0$ or $1$ without considering negative samples. 
On the other hand, when using only $\mathcal{L}_{neg}$, the model shows higher frame-level accuracy and lower F1-scores. This is because it helps the model predict all frames as $0$ for negative pairs, thereby predicting the alignment span more tightly. Ultimately, the combination allows the model to benefit from the strengths of each loss component, resulting in improved frame-level accuracy and F1-scores.

\begin{table}[t!]
  \renewcommand{\arraystretch}{1.2}
  \renewcommand{\tabcolsep}{1.2mm}
  \centering
  \vspace{-1mm}
  \caption{\textbf{Ablation study on self-training.} The performance improvement from self-training over 2 epochs is marginal. This demonstrates that self-training for just 1 epoch is sufficient to achieve satisfactory performance.}
  \resizebox{0.999\linewidth}{!}{
  \begin{tabular}{l|cccc}
    \toprule
    Method & frame-acc ($\%$) $\uparrow$ & F1@.10 $\uparrow$ & F1@.25 $\uparrow$ & F1@.50 $\uparrow$ \\
    \midrule
    Ours (self-training for 1 epoch)& 77.22 & 81.39 & 75.03 & 63.81 \\
    Ours (self-training for 2 epoch) & 77.39 & 81.27 & 75.19 & 63.96 \\
    \bottomrule
  \end{tabular}}
  \label{tab:selftraining}
  \vspace{-3mm}
\end{table}

\newpara{Ablation study on self-training.}
As shown in~\Tref{tab:selftraining}, we conduct another round of self-training, resulting in marginal performance improvements across most metrics. This outcome underscores that even if the self-training process is performed only once, it shows sufficiently saturated performance.

\subsection{Qualitative results} \label{sec:qualitative}
In this subsection, we qualitatively analyse frame-level model predictions to verify the feasibility of the proposed method. To do this, we visualise the predicted timeline and alignment probability in~\Fref{fig:qualitative}. We sequentially show the output for the shifted audio subtitle, the SAT model, our model without selective alignment (denoted as \emph{Ours w/o SA loss}), and our full model, including ground truth timing.

\newpara{Predicted alignment timeline.} 
In the upper part of~\Fref{fig:qualitative}, 
while the shifted subtitle prior provides an approximate position, it is largely misaligned with the ground truth. 
The SAT model exhibits a stronger bias towards the prior alignment, failing to align well with the text query. 
In contrast, both our model without selective loss and the full model show improved alignment over the SAT model. 
This improvement stems from a better understanding of the relationship between text queries and video content, particularly in sign language linguistics.
Upon closer inspection, applying selective alignment loss further improves alignment with the ground truth. This is attributed to the model's ability to fine-tune alignment by reducing attention to frames unrelated to the text query.

\newpara{Alignment probability visualisation.}
In the lower part of~\Fref{fig:qualitative}, we further visualise the alignment probabilities to analyse where the models focus on. The SAT model tends to emphasise a timeline biased towards the input prior. 
The model trained without selective alignment loss exhibits better alignment performance than the SAT model but shows low confidence for some of the positive frames.
Finally, our full model showcases reduced temporal attention for frames that do not correspond to the text query, while increasing attention for matching segments. This results in predictions that closely align with the ground truth, indicating the effectiveness of the selective alignment loss. This loss facilitates a better understanding of the correlation between sign language and text, particularly in accurately aligning video segments with corresponding textual content.

\section{Conclusion}
In conclusion, this work presents a novel framework aimed at enhancing the accuracy of sign annotations compared to existing baselines. 
we address the oversight in previous research regarding the distinct grammatical disparities between sign and spoken languages. 
Moreover, we tackle the challenges posed by noisy and weak supervision inherent in datasets collected from TV broadcasts. 
To mitigate noisy supervision, we introduce a selective alignment loss mechanism, penalising misalignments unrelated to text queries and ensuring precise alignment with subtitles. Additionally, we alleviate weak supervision by implementing a self-training strategy, which leverages pseudo-labels generated by our model to further enhance performance. 
Our findings pave the way for future research in sign language processing and interpretation, ultimately contributing to improved accessibility and inclusivity for the deaf and hard-of-hearing community.


\bibliographystyle{IEEEtran}
\bibliography{egbib.bib}

\end{document}


\title{Deep Understanding of British Sign Language\\for Sign to Subtitle Alignment}

\maketitle
asdfasdfasdfasdf

